\documentclass[10pt,letterpaper]{article}
 
\usepackage{cogsci}
 
\cogscifinalcopy 
 
\usepackage{pslatex}
\usepackage{apacite}
\usepackage{graphicx}
\usepackage{caption}
\usepackage{subcaption}
 
\title{Semantic features of object concepts generated with GPT-3}
 
 \author{{\large \bf Hannes Hansen (hansen@cbs.mpg.de)} \\
  Vision and Computational Cognition Group, Max Planck Institute for Human Cognitive and Brain Sciences, Stephanstraße 1a \\
  04103 Leipzig Germany
  \AND {\large \bf Martin N. Hebart (hebart@cbs.mpg.de)} \\
  Vision and Computational Cognition Group, Max Planck Institute for Human Cognitive and Brain Sciences, Stephanstraße 1a \\
  04103 Leipzig Germany}

\begin{document}
 
\maketitle

\begin{abstract}
Semantic features have been playing a central role in investigating the nature of our conceptual representations. Yet the enormous time and effort required to empirically sample and norm features from human raters has restricted their use to a limited set of manually curated concepts. Given recent promising developments with transformer-based language models, here we asked whether it was possible to use such models to automatically generate meaningful lists of properties for arbitrary object concepts and whether these models would produce features similar to those found in humans. To this end, we probed a GPT-3 model to generate semantic features for 1,854 objects and compared automatically-generated features to existing human feature norms. GPT-3 generated many more features than humans, yet showed a similar distribution in the types of generated features. Generated feature norms rivaled human norms in predicting similarity, relatedness, and category membership, while variance partitioning demonstrated that these predictions were driven by similar variance in humans and GPT-3. Together, these results highlight the potential of large language models to capture important facets of human knowledge and yield a new approach for automatically generating interpretable feature sets, thus drastically expanding the potential use of semantic features in psychological and linguistic studies.
 
\textbf{Keywords:}
semantic features; conceptual knowledge; natural language processing; GPT-3
\end{abstract}

\section{Introduction}
A central aim in the cognitive sciences is to understand the nature of human conceptual knowledge. This knowledge is often characterized through semantic features, which form a set of minimal semantic descriptions of concepts and which have been at the heart of much theorizing about categorization (\citeNP{nosofsky1986attention}; \citeNP{rosch1973internal}), semantic memory \cite{murphy2004big}, and semantic processing more generally \cite{cree2003analyzing}. For example, the concept \emph{car} can, among others, be described by the features \emph{is a vehicle} and \emph{has four wheels}. The relationship of this concept to other concepts can then be quantified by evaluating the similarities and differences to the features of other concepts.\newline  
A common approach for attaining semantic features of concepts is to instruct humans to list properties for a given concept, for example by asking \emph{what are the properties of a cow?}. The popularity of such empirically-generated semantic features has led researchers to create semantic feature production norms for a larger number of concepts (\citeNP{devereux2014centre}; \citeNP{mcrae2005semantic}), which have been invaluable for improving our understanding of semantic representations. At the same time, the impact of these norms has remained constrained to the set of concepts and features that have been made publicly available. Creating new norms requires collecting responses from hundreds of participants and necessitates extensive manual curation by researchers, inherently restricting the scope of such approaches. If there was a computational model that contained the knowledge sufficient for generating feature norms of similar quality to humans, this would drastically expand the possible scope of research with semantic features in the study of conceptual knowledge.\newline
In recent years, there have been strong advances in the field of natural language processing. So-called transformer models, such as BERT \cite{devlin2018bert} or GPT-3 \cite{brown2020language}, often approach human-level performance in diverse language understanding tasks (\citeNP{floridi2020gpt}; \citeNP{wang-etal-2018-glue}) and can even be used to produce prose that can be difficult to distinguish from human-generated text \cite{dale2021gpt}. While the general linguistic understanding and reasoning ability of these models are still far from perfect \cite{marcus2020next}, they may offer a valuable computational tool for automatically producing semantic features for an arbitrary number of concepts \cite{derby2019feature2vec}, thus leveraging the statistical structure of knowledge present in their immense training text corpora for addressing central questions in semantic cognition research \cite{bhatia2021transformer}.\newline
Here we tested the degree to which recent transformer models can be used for automatic production of semantic features and whether the produced features mirror those found in humans. To this end, we used GPT-3 to generate a semantic feature norm for 1,854 diverse concepts of concrete objects. We chose concrete objects for two reasons. First, concrete objects have been used in much research on conceptual representations and are indeed used in several existing feature production norms, thus providing a valuable human baseline to relate our results to. Second, similarity ratings for these 1,854 objects have recently become available \cite{hebart2020revealing}, providing a broad test case for validating these features with existing similarity ratings.
 
\begin{figure*}[h]
\centering
\includegraphics[width=1\textwidth]{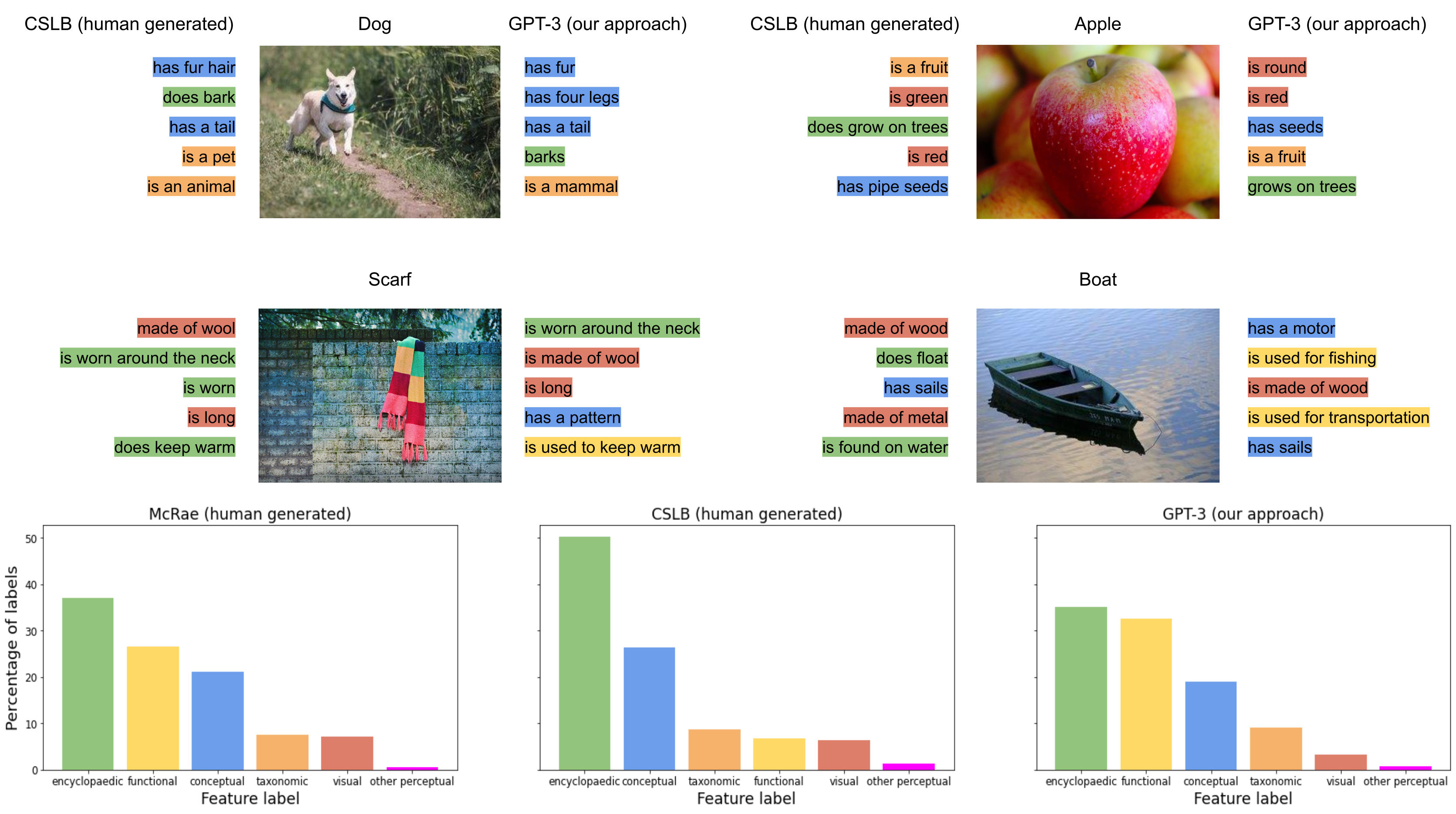}
\caption{Comparison of features from the CSLB norms and features generated using GPT-3, including feature examples (top) and feature label distributions (bottom). Please note that images were not part of the study and are used for illustration only. Bootstrapped confidence intervals for feature label distributions were too small to be displayed.}
\label{labels}
\end{figure*}
 
\begin{table*}[!ht]
\caption{Descriptive statistics of human and GPT-3 generated feature norms}
\begin{center}
\begin{tabular}{lllll}
\hline
 Feature   & CSLB & McRae & GPT-3 (preprocessed) & GPT-3 (preprocessed + filtered)\\
\hline
Number of concepts   &   638  & 541  & 1,854 & 1,854 \\
Total number of features        &  22,667 & 7,259  & 189,126 & 124,569 \\
Number of unique features     & 5,929  & 2,524 & 63,467 & 11,683  \\
Number of features per concept       &   35.52  & 13.42  & 102.06 & 67.35 \\
Share of unique features to all features & 26.16\% & 34.77\% & 33.55\% & 9.37\% \\
\hline
\end{tabular}
\label{sample-table}
\end{center}
\end{table*}
 
\section{Related research}
Several previous studies have investigated the use of corpus-based language models to produce sets of features mirroring human conceptual knowledge. Static word embeddings, including \emph{word2vec} \cite{mikolov2013efficient} or \emph{GloVe} \cite{pennington2014glove} are trained on lexical co-occurrences in large text corpora and provide decent predictions of human similarity ratings \cite{hill2015simlex}. However, the features of these models typically lack interpretability \cite{subramanian2018spine}. \citeA{rubinstein2015well} used word embeddings to directly predict a small set of semantic features, concluding that distributed language models may be better at capturing taxonomic than attributive features. \citeA{psrl} used partial least squares regression to map word embeddings to feature norms, with a more recent approach using a non-linear mapping based on a multilayer perceptron \cite{li2019mapping}. \citeA{derby2019feature2vec} proposed \emph{Feature2Vec}, a method which combines the information from word embeddings with human feature norms by projecting the features into the word embedding space. Despite good overall performance, these methods rely on a fixed feature vocabulary, making it only possible to generate features for new concepts but not completely new features.\newline
More recently, \citeA{bhatia2021transformer} investigated the use of transformer models to model human conceptual knowledge by finetuning a pretrained BERT model \cite{devlin2018bert} on a broad set of features and probing the model whether the concept-feature pairs were correct. The model correctly predicted concept features even outside of the training set with good accuracy, providing an important step towards general purpose feature generators. However, this method still requires probing existing feature-concept pairs, rather than generating new features. Thus, it remains unknown to what degree it is possible to generate arbitrary features for arbitrary concepts. Finally, \citeA{bosselut-etal-2019-comet} and \citeA{petroni-etal-2019-language} proposed models of commonsense knowledge based on GPT and BERT. Models are finetuned on subject-relation-object triplets, with the task of predicting the object (e.g. \emph{mango-IsA-fruit}). While these models can partially generate features for concepts, they are limited in that they require a prori specification of relations. 
 
\begin{figure*}[h]
\centering
\includegraphics[width=\textwidth]{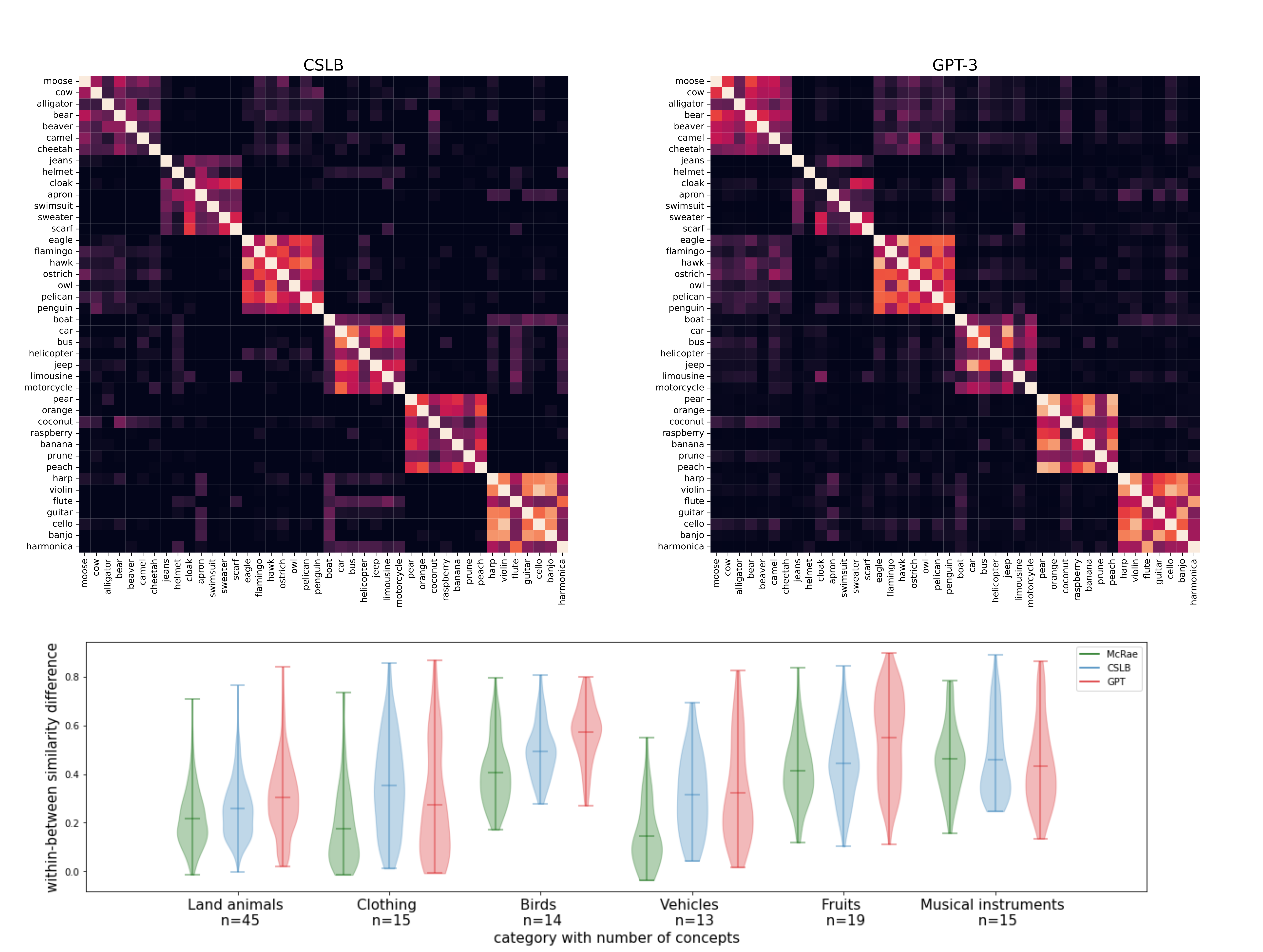}
\caption{Representational similarity matrices for 6 concepts in the categories land animals, fruits, vehicles, fruits, birds and clothing using the CSLB and GPT-3 feature norms (top) and pairwise cosine similarities per category (bottom).}
\label{rsa}
\end{figure*}
 
\section{Methods}
\subsection{Feature collection from GPT-3}
We probed GPT-3 (DaVinci version) to generate semantic features for 1,854 object concepts from the THINGS database \cite{hebart2019things}, using a text completion task. To instruct GPT-3 on this task, we presented it with a question about an object (e.g. \emph{What are the properties of a chair?}), had it generate an answer, and replaced this answer with features from the McRae feature norm (\citeNP{mcrae2005semantic}) which served as ground truth (e.g. \emph{It is furniture, it is made of wood, etc.}). When continuing this process, generated features appeared to no longer improve in quality after three question and answer sequences, so we chose this number as a trade-off between monetary cost and performance. Once GPT-3 was primed on the task, this was followed by an open question for each of the 1,854 object concepts, after which the answer was collected and both the answer and the question deleted to keep the context static across all trials. To reduce bias induced by specific concepts and associated features and to better mirror experimental approaches that merge data across human participants, we collected 30 runs, each time using a different set of example concepts. In cases where a concept occurred multiple times with different meanings (e.g. bat as animal or bat as sports item), we added a superordinate category to the training example in parantheses.
 
\begin{figure*}[h]
\centering
\includegraphics[width=1\textwidth]{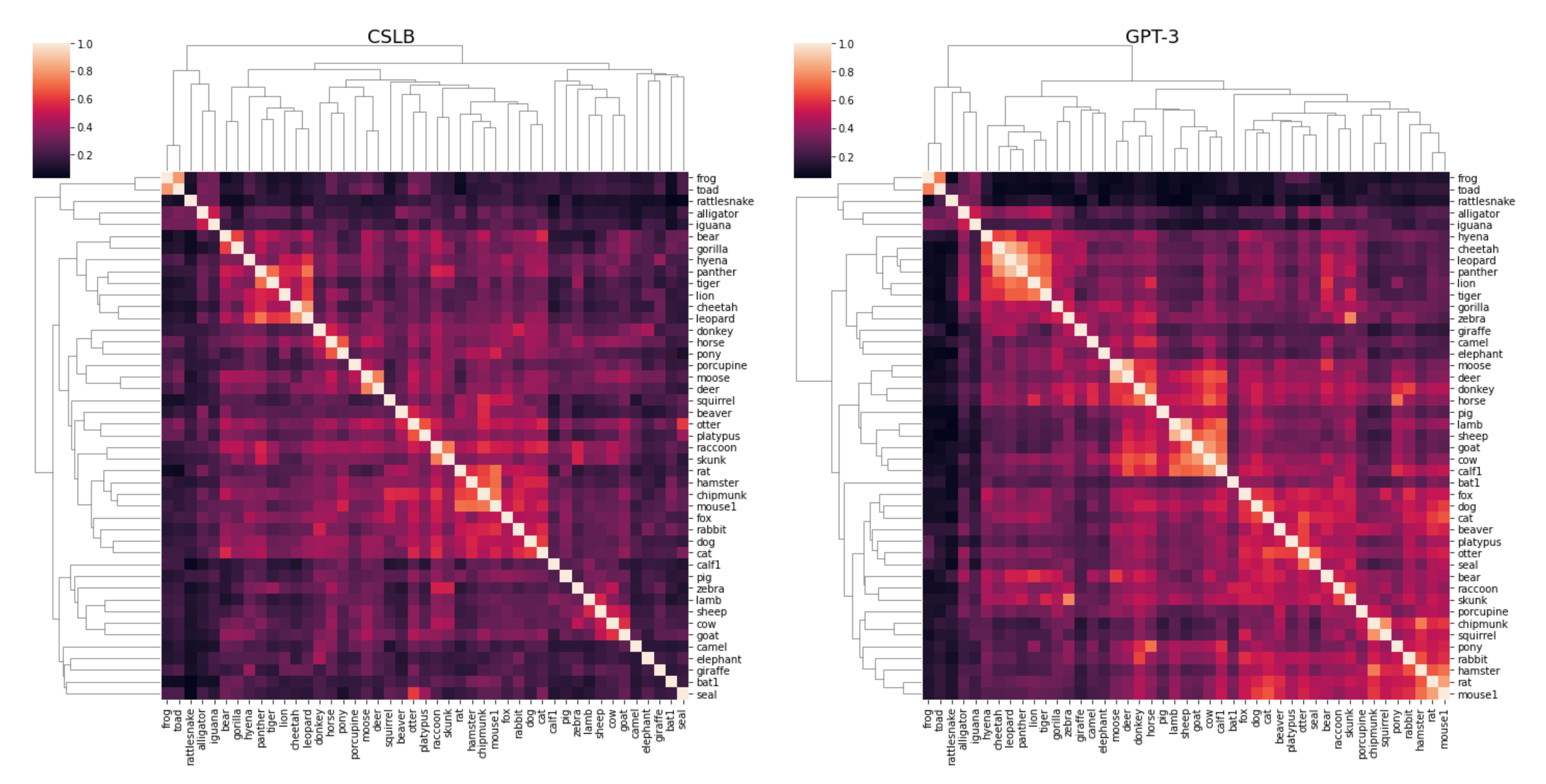}
\caption{Hierarchical clustering for 45 land-animal concepts using the CSLB and GPT-3 feature norms, highlighting similarities and differences for within category representational structure.}
\label{cluster}
\end{figure*}
 
\subsection{Preprocessing of features}
For better comparability to humans, we preprocessed and normed generated features. We automated preprocessing by using part of speech tagging with the Python library \emph{spacy}\footnote{https://spacy.io/} and applying a set of preprocessing rules (see below).\newline
The produced answers consisted of lists of features which were split at commas in order to attain raw features. Next, raw features that were classified as nonsensical, consisted of a single word, or were tautological (e.g. \emph{a rose is a rose}) were removed. In addition, features not beginning with a pronoun (e.g. \emph{green color} rather than \emph{it has green color}) or features containing non-ASCII characters (\emph{it has ©}) or question marks were removed. Finally, qualifier adverbs (e.g. \emph{usually} or \emph{really}) were removed, in line with previous approaches (\citeNP{devereux2014centre}). This affected 0.75\% of all features.\newline
Next, long features with a subordinate clause (\emph{which, that, when, if, but}) were shortened by removing the subordinate part. For example, a feature like \emph{it is a car that drives} was shortened to \emph{it is a car}. This affected 1.83\% of all features.\newline
These clean and concise features were furthermore normed. Nested features, containing multiple units of information, were extracted. For example, a feature like \emph{it is a big tree} was decomposed automatically into \emph{it is a tree} and \emph{it is big}. A feature like \emph{it is blue and green} was decomposed into \emph{it is blue} and \emph{it is green}. Next, features containing synonyms, e.g. \emph{it is a car} and \emph{it is an automobile} were collapsed using Wordnet synsets in the Python library \emph{nltk}\footnote{https://www.nltk.org/} by choosing the more frequent synset. However, to avoid merging non-synonymous words, two words were only considered as synonyms if their most frequent synset was the same. 4.3\% of all features were replaced.
 
\subsection{Filtering}
The generated feature norm partially consisted of overly sparse features, with many features that were unique to individual concepts. However, a smaller set of features is desirable both for reasons of better interpretability and for reduced computational cost. Therefore, we removed features that occurred very infrequently within each concept. This also made the feature norm more comparable in size to human generated norms. To identify a cutoff, we plotted the number of unique features after removal of infrequent features and chose the elbow point at k=4. Importantly, while this step strongly reduced the number of unique features (see Table \ref{sample-table}), it did not affect performance in validating the norm.
 
\section{Results}
\subsection{Comparison with human feature norms}
As a first analysis, we compared the GPT-3 feature norms with human feature norms McRae \cite{mcrae2005semantic} and CSLB \cite{devereux2014centre}, using descriptive statistics. We did not use the Buchanan norm \cite{buchanan2019english} as it was composed only of associative features. As seen in Table \ref{sample-table}, the preprocessed GPT-3 feature norm was computed for a larger number of concepts, leading to a larger number of total features and more unique features. Without filtering, a similar share of unique features can be seen as in the McRae norm, indicating a similar level of redundancy of features across concepts as found in humans. After excluding rare features, only around 9\% of all features remained unique. Of note, GPT-3 produced a much larger number of features per concept, indicating that human feature production may be limited to a sparser set of features than those found in GPT-3.
 
\subsection{Label distribution}
Figure \ref{labels} (top) shows several example concepts with their five most frequent features. The results indicate many similarities and no obvious errors in the types of labels assigned by GPT-3. To directly compare the distribution of semantic features between humans and GPT-3, samples of features from the CSLB feature norms and GPT-3 were labeled into the categories taxonomic, visual perceptual, other perceptual, conceptual, functional and encyclopedic. We use a slightly different naming scheme of feature types to McRae and CSLB to allow for more fine-grained differences between conceptual, functional, and encyclopedic features. To estimate the distribution of features in all three norms, we randomly sampled 500 features from the 317 concepts shared between CSLB, McRae, and GPT-3 and 500 features from concepts outside of the intersection. The corresponding feature labels were assigned manually.\newline
The results in Figure \ref{labels} (bottom) show that humans and GPT-3 mostly rely on encyclopedic features and less on perceptual features. GPT-3 contained a larger number of functional features as compared to the CSLB norm. However, the differences in the distribution to the McRae norm, which GPT-3 had been trained on, were less prominent, indicating that differences of GPT-3 to CSLB may be driven more by differences in populations for the creation of human norms or norming processes, rather than intrinsic differences in representations.  Overall, this analysis yielded no obvious differences to human norms in the types of semantic features that were generated.
 
\begin{table}[!ht]
\caption{Similarity and relatedness prediction}
\label{correlation}
\centering
\begin{tabular}{llll}
\hline
Pearson correlation    & McRae & CSLB & GPT-3 \\
\hline
MEN (n=55 word pairs) &  0.77 & 0.77 & 0.79 \\
Simlex-999 (n=26 word pairs) & 0.60 & 0.63 & 0.62 \\
THINGS (n=317 concepts) & 0.56 & 0.63 & 0.62 \\
\hline
\end{tabular}
\end{table}
 
\subsection{Prediction of category structure based on feature similarity}
Next we tested the degree to which GPT-3 generated feature norms produced reasonable category structure and how they compared to existing norms. For better comparability, several analyses were conducted in a similar fashion to those found for the creation of the CSLB norm \cite{devereux2014centre}. For a fair comparison, we restricted our analyses to the 317 concepts shared between the McRae, CSLB and GPT-3 feature norms. Similarities were based on the cosine similarity of the concepts across feature vectors composed of the production frequency per feature, for human norms across participants and for GPT-3 across the 30 runs for which it had been primed with different examples.\newline
First, we inspected the superordinate category structure. Mirroring the results of \cite{devereux2014centre}, we based these analyses on six exemplars of the six categories \emph{land animals}, \emph{clothing}, \emph{birds}, \emph{vehicles}, \emph{fruits}, and \emph{musical instruments}. If the produced features prove to be useful for predicting high-level category structure, we would expect consistently high within-category similarity and low between-category similarity. A visualization of the similarity structure of the CSLB norm with the GPT-3 generated norm is shown in Figure \ref{rsa} (top). As is evident from the figure, both norms show excellent category structure, clearly distinguishing high-level categories from each other. To quantify similarities and differences between norms, we expanded the set of concepts available within each superordinate category to the 317 concepts and computed the mean within-category similarity minus the mean between-category similarity of each concept. The results are shown in Figure \ref{rsa} (bottom). Overall, GPT-3 showed comparable or clearer category structure than human norms, with the only exception of \emph{clothing}. Together, these results demonstrate that high-level category structure can be extracted from GPT-3 generated feature norms, with similar performance to humans. \newline
Beyond between-category effects, we explored whether the within-category similarity structure was meaningful in GPT-3 generated norms. To this end, we performed hierarchical clustering of the 45 land animal concepts. Overall, we expected high similarities between all animals but also more fine-grained differences. The results comparing CSLB with GPT-3 norms are shown in Figure \ref{cluster}. Overall, the similarities between animals were higher in the GPT-3 norm than CSLB. A clear clustering of highly similar animals is found in both matrices (e.g. \emph{lamb} and \emph{sheep}). However, the subordinate category structure was slightly different between GPT-3 than CSLB. For example, farm animals clustered in GPT-3 but were more distributed in CSLB, while dangerous animals clustered more closely in CSLB. In sum, while the overall category structure was similar, there were fine-grained differences in the representations derived from semantic features through GPT-3 as compared to the CSLB norm.
 
\begin{figure*}[h]
\centering
\includegraphics[width=0.9\textwidth]{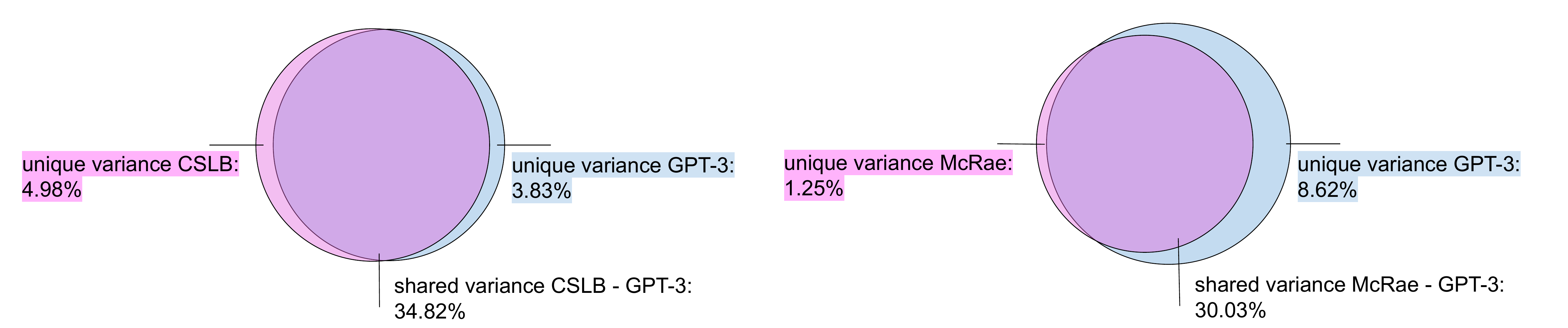}
\caption{Variance partitioning of human similarity judgements, demonstrating strong overlap between human generated and GPT-3 generated norms in predicting human similarity judgments.}
\label{variance}
\end{figure*}
 
\subsection{Prediction of similarity and relatedness ratings}
Predictions of similarity and relatedness tasks are often seen as a gold standard for evaluating the relationship between semantic features and our conceptual representations. To identify the degree to which GPT-3 generated norms could be used to predict similarity and relatedness ratings, we used the overlap between McRae, CSLB, and GPT-3 generated norms with two existing datasets commonly used as natural language processing benchmarks: The MEN dataset \cite{bruni2014multimodal} was used for word relatedness and the Simlex-999 dataset \cite{hill2015simlex} for word similarity. We intended to include other common benchmark datasets but did not find sufficient overlap in concepts. Beyond these datasets, we used human similarity scores from THINGS \cite{hebart2020revealing}, which were available for all included objects. Similarities were compared by using the Pearson correlation of the flattened lower triangular part of the similarity matrices.\newline
The overlap in the number of concepts as well as the performance for McRae, CSLB, and GPT-3 generated norms are found in Table \ref{correlation}. While CSLB performed better than McRae for the prediction of THINGS, across all three datasets, the performance of CSLB and GPT-3 was comparable. This indicates that the representations derived from GPT-3 generated norms were of similar quality as those found in human norms.\newline
What is left open by these correlations is whether the predictions of similarity were based on similar information in McRae, CSLB, and GPT-3, or whether GPT-3 had access to other information not reported by humans. To address this question, we carried out variance partitioning, identifying the unique and shared variance components explained in the THINGS similarity dataset. The results of this analysis are shown in Figure \ref{variance}. Overall, there was strong overlap in the explained variance between McRae and GPT-3 as well as CSLB and GPT-3, with GPT-3 subsuming much of the variance explained by McRae, while explaining very similar portions of variance than CSLB. Overall, this result demonstrates that, indeed, GPT-3 is relying on similar information as CSLB for predicting human similarity.
 
\section{Discussion}
Overall, the results demonstrate that GPT-3 can be used for automatically generating semantic feature norms for a large number of concrete objects. Frequently produced features were meaningful and comparable in distribution to those found in humans. Further, the results showed that superordinate categories were well identified by the resulting features and that within-category structure was reasonable. Predictions of similarity and relatedness were comparable to humans and relied on similar information. Overall, this demonstrates that GPT-3 can serve as an effective automatic feature generator, opening up an efficient and rapid approach to generate large numbers of features for diverse concepts.\newline
Of note, there were some differences to humans. Overall, GPT-3 produced a much larger number of features, yet reducing this set through filtering led to very similar results in the prediction of similarity ratings. Further, the fine-grained similarity structure was slightly different to CSLB. Future work is needed to investigate these differences and the degree to which they are related to differences between representations in humans and GPT-3, or whether they were driven by the norming process itself.\newline
GPT-3 was primed using only three examples, which highlights the simplicity of our proposed approach but which may also introduce bias. Rather than reflecting the knowledge available to the model, it may in fact mimic the process of feature generation produced by humans in the three examples it was provided. Other, more effective priming procedures may be discovered in the future that constitute less bias and are better at revealing the knowledge available in such models. Future work may also investigate the influence of model complexity on feature generation. \newline
Nevertheless, the fact that GPT-3 and humans relied on similar information for predicting similarity ratings is relevant in its own respect, indicating that GPT-3 may indeed contain a lot of information useful for modeling important aspects of conceptual knowledge \cite{bhatia2021transformer}. As a consequence, it may be possible to use GPT-3 to generate other types of norms, for example ratings of animacy, graspability, or size \cite{grand2022semantic}. We did not test whether GPT-3 was able to produce meaningful features for more abstract concepts or verbs, which is an important avenue for future research. While better computational models for producing semantic features may exist, our work demonstrates that it is already possible to create semantic features for concrete concepts with human level performance in predicting similarity ratings using GPT-3.
 
\section{Conclusions}
Here, we introduced a new approach for automatically generating semantic features for diverse concepts using the recent transformer-based model architecture GPT-3. Our results demonstrate that recent large language models are indeed able to accurately reflect important aspects of human conceptual knowledge. The approach opens the door to automatic feature generation for arbitrary concepts, thus widening the potential scope of semantic features for research in psychology and linguistics. To promote the general use of this method and results, the GPT-3 generated raw data as well as the final feature norm of the 1,854 object, including the code to probe GPT-3 and to preprocess and filter raw features, are made publicly available\footnote{https://github.com/ViCCo-Group/semantic\_features\_gpt\_3}.
 
\bibliographystyle{apacite}
 
\setlength{\bibleftmargin}{.125in}
\setlength{\bibindent}{-\bibleftmargin}
 
\bibliography{CogSci_Template}

\end{document}